\documentclass[prl,twocolumn,twoside,showpacs,byrevtex,superscriptaddress]{revtex4}

\usepackage{currfile}
\usepackage{graphicx,epsfig,verbatim,enumerate}
\usepackage{amssymb,amsmath,mathdots,newfloat}
\usepackage{ifthen,caption}

\usepackage{lipsum}
\usepackage[framemethod=TikZ]{mdframed}

\usepackage{tikz-cd}

\usepackage{hyperref}
\hypersetup{
        colorlinks = true,
        allcolors = blue
}
\usetikzlibrary{arrows}
\tikzset{
  commutative diagrams/.cd,
  arrow style=tikz,
  diagrams={>=space}}

\newboolean{twocolswitch}

\DeclareFloatingEnvironment[fileext=examp,placement={!ht},name=Example]{exampl}

\newenvironment{examp}[1]{
    \begin{exampl}[t!]
    \vspace{8pt}
    \begin{mdframed}[roundcorner=10pt,backgroundcolor=gray!10]
    \vspace{7pt}    
}
{
    \vspace{15pt}
    \end{mdframed}    
    \end{exampl}
}

\newcommand{\sindex}[1]{}
\newcommand{\nindex}[1]{}

\newcommand{\www}[1]{\url{#1}}

\usepackage{lettrine}

\usepackage{color}

\newcommand{\simonalpha}{\theta}

\newcommand{\numberoftexts}{k}

\newcommand{\Nav}{N_\text{avg}}
\newcommand{\Nma}{N_\text{max}}
\newcommand{\Nmi}{N_\text{min}}
\newcommand{\memnum}{n}
\newcommand{\memlessnum}{N}

\RequirePackage{color}\definecolor{RED}{rgb}{1,0,0}\definecolor{BLUE}{rgb}{0,0,1} 

\setboolean{twocolswitch}{true}

\begin{document}

\title{

Text mixing shapes the anatomy of rank-frequency distributions: \\
A modern Zipfian mechanics for natural language
}

\author{
\firstname{Jake Ryland}
\surname{Williams}
}

\email{jake.williams@uvm.edu}

\affiliation{Department of Mathematics \& Statistics,
  Vermont Complex Systems Center,
  Computational Story Lab,
  \& the Vermont Advanced Computing Core,
  The University of Vermont,
  Burlington, VT 05401.}

\author{
\firstname{James P.}
\surname{Bagrow}
}
\email{james.bagrow@uvm.edu}

\affiliation{Department of Mathematics \& Statistics,
  Vermont Complex Systems Center,
  Computational Story Lab,
  \& the Vermont Advanced Computing Core,
  The University of Vermont,
  Burlington, VT 05401.}

\author{
\firstname{Christopher M.}
\surname{Danforth}
}
\email{chris.danforth@uvm.edu}

\affiliation{Department of Mathematics \& Statistics,
  Vermont Complex Systems Center,
  Computational Story Lab,
  \& the Vermont Advanced Computing Core,
  The University of Vermont,
  Burlington, VT 05401.}

\author{
\firstname{Peter Sheridan}
\surname{Dodds}
}
\email{peter.dodds@uvm.edu}

\affiliation{Department of Mathematics \& Statistics,
  Vermont Complex Systems Center,
  Computational Story Lab,
  \& the Vermont Advanced Computing Core,
  The University of Vermont,
  Burlington, VT 05401.}

\begin{abstract}
  Natural languages are full of rules and exceptions. 
One of the most famous quantitative rules is Zipf's law
which states that the frequency of occurrence of a word is approximately inversely proportional to its rank. 
Though this `law' of ranks has been found to hold across disparate texts and forms of data, 
analyses of increasingly large corpora over the last 15 years 
have revealed the existence of two scaling regimes.
These regimes have thus far been explained by a hypothesis suggesting 
a separability of languages into core and non-core lexica. 
Here, we present and defend an alternative hypothesis,
that the two scaling regimes result from the act of aggregating texts.
We observe that text mixing leads to an effective decay of word introduction,
which we show provides accurate predictions of the location and severity of breaks in scaling.
Upon examining large corpora from 10 languages 
in the Project Gutenberg eBooks collection (eBooks),
we find emphatic empirical support for the universality of our claim.

\end{abstract}

\pacs{89.65.-s,89.75.Da,89.75.Fb,89.75.-k}

\maketitle

\section{Zipf's law and (non) universality}

Given some collection of distinct kinds of objects
occurring with frequency $f$ and associated rank $r$ according
to decreasing frequency, Zipf's law is said to be fulfilled 
when ranks and frequencies are approximately inversely proportional:
\begin{equation}
   f(r)
   \sim 
   r^{-\simonalpha},
\end{equation}
typically with $\simonalpha \simeq 1$.
Though Zipf's functional form has been found to be a reasonable one for disparate forms of 
data, ranging from frequencies of words to sizes of cities in Zipf's original work \cite{zipf1935a,zipf1949a},
its lack of \textit{total} universality in application to natural languages is now widely acknowledged
\cite{cancho2001a,montemurro2001a,gerlach2013a,kwapien2010a,petersen2012a,williams2014a}.

Recently it was suggested~\cite{cancho2001a,montemurro2001a}
that large corpora exhibit two scaling regimes (delineated by some $b>0$):
\begin{equation}
  f(r) \sim \left\{
  \begin{array}{lr}
    r^{-\simonalpha}, & : r \leq b\\
    r^{-\gamma}, & : r > b
  \end{array}
  \right.,
\end{equation}
the first being that of Zipf ($\simonalpha=1$) and the second distinctly more variable~\cite{montemurro2001a},
(though generally $\gamma > 1$). Ferrer and Sol\'e hypothesized in~\cite{cancho2001a} 
that these two regimes reflected a division of natural languages into two lexical subsets---the 
kernel (core) and unlimited (non-core) lexica. 

We observe that in all studies finding dual scalings that the texts analyzed are of mixed origin, 
that is, they are not derived from a single author, or even a single topic. 
Montemurro indicated in~\cite{montemurro2001a}
that combining heterogeneous texts could generate effects that shield investigators from the true
underlying nature of this second scaling regime:
\begin{quote}
  To resolve the behavior of those [high rank] words we need a significant increase in volume of 
  data, probably exceeding the length of any conceivable single text. Still, at the same time it is 
  desirable to maintain as high a degree of homogeneity in the texts as possible, in the hope of 
  revealing a more complex phenomenology than that simply originating from a bulk average of a wide 
  range of disparate sources.
\end{quote}
With this inspiration, we focus on understanding the effects of combining texts of varying 
heterogeneity---a process we refer to as ``text mixing''.

\section{Stochastic models}

In the years following Zipf's original work, various stochastic models have been proposed for 
the generation of natural language vocabularies. The first of these was that proposed by Simon 
\cite{simon1955a}, and based on Yule's model of evolution~\cite{yule1924a}. This work is a powerful 
companion to understanding Zipf's empirical work, and can be seen as the natural antecedent of 
the rich-gets-richer models~\cite{barabasi1999a,krapivsky2001a} for growing networks that 
have interested the complex systems community over recent years. Indeed, perhaps the most important 
piece we may draw from Simon's model is that a rich-gets-richer mechanism is a reasonable 
one for the growth of a vocabulary.

An important limitation of Simon's model is that it is only capable of producing a single scaling 
regime, which, as we know is an incomplete picture. Furthermore, the scalings accessible via the 
Simon model were strictly less severe than the `universal' $\simonalpha=1$ exponent. So, if one assumes the 
Simon model as truth, with a fixed word introduction rate $\alpha_0$, Zipf's exponent should be 
variable and necessarily less than $1$, though empirically found indistinguishable from $1$, 
that is $\simonalpha = 1 - \alpha_0$, with $\alpha_0 \ll 1$~\cite{simon1955a}.

Recently, a modification to Simon's model was proposed in which two types of words could be produced---core 
and non-core words~\cite{gerlach2013a}. As a built-in feature of the core/non-core vocabulary (CNCV) model, 
the size of the core set of words was prescribed to be finite, while the non-core was allowed to expand 
indefinitely. Aside from introducing two classes of words, the most important distinction of this model from 
its predecessor was a rule for the decay in the rate of introduction of new words, $\alpha$. Along with 
producing the CNCV model they showed that when $\alpha$ decays as a power-law with exponent $-\mu$, of the 
number of unique words, $\memnum$, the relationship between $\mu$ and the lower rank-frequency exponent, $\gamma$, 
is a difference of $\simonalpha$, i.e., 
\begin{equation}
  \alpha(\memnum)
  =
  \alpha_0\cdot\memnum^{-\mu}
  \Rightarrow
  f(r)
  \sim
  r^{-(\simonalpha+\mu)},
\end{equation}
with $\gamma = \simonalpha+\mu$~\cite{gerlach2013a}.
The distinction between word types provided a means for postponing the point at 
which their power law decay would occur, thereby generating two scaling regimes. 
We note that the severity of the second scaling was \textit{only} contingent upon the existence of a 
decay in the rate of introduction of new words, and that this decay was imposed, 
rather than the result of the existence of two word types.
We are therefore drawn to find an explicit mechanism capable of producing power-law decaying word introduction rates, 
and hence multiple scaling regimes.

\section{Text mixing}

\begin{figure}[t!]
  \centering
  \includegraphics[width=0.475\textwidth, angle=0]{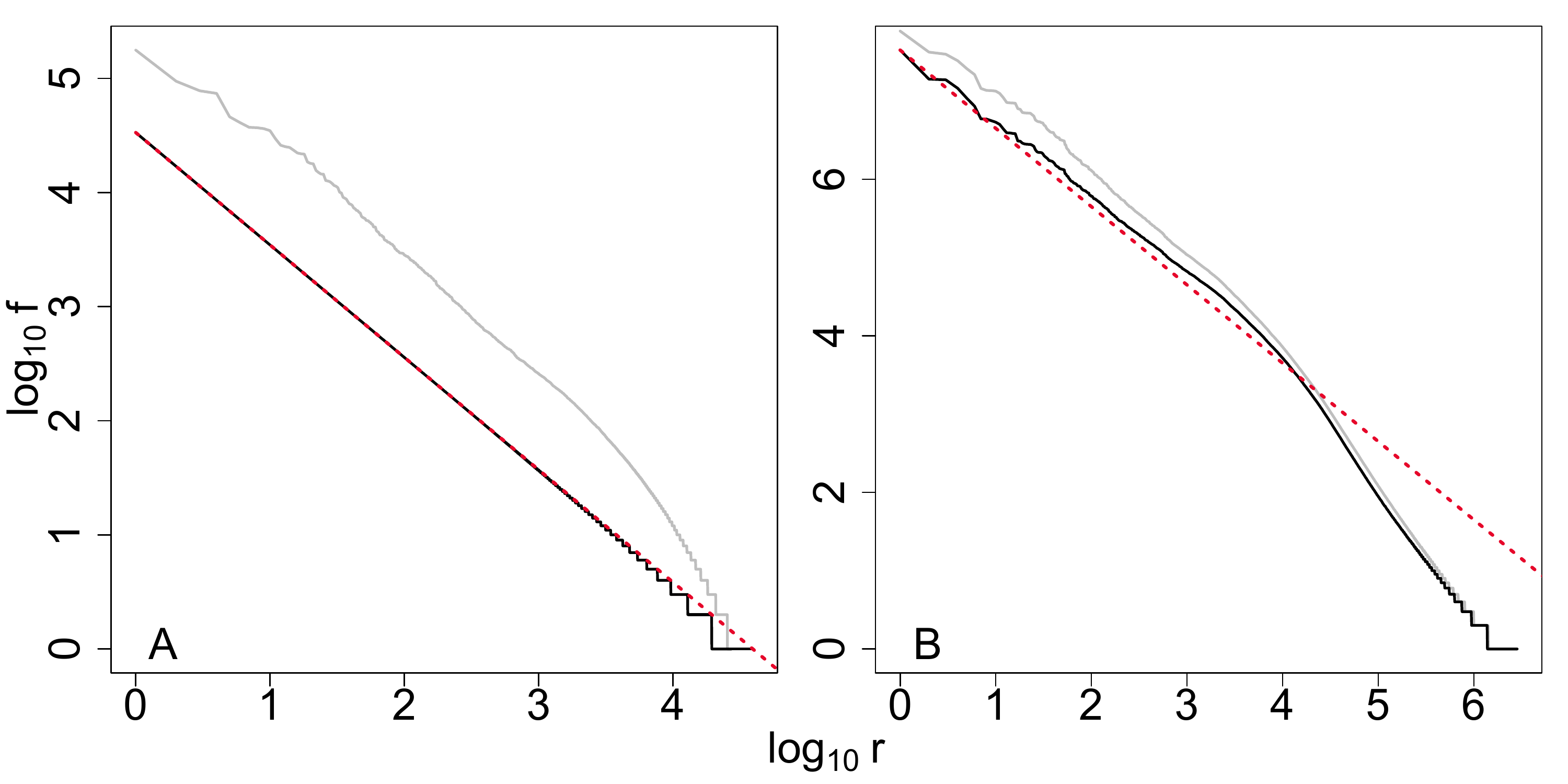}
  \caption{
        (\textbf{A}) An idealization (black points) of a rank-frequency distribution (gray points)
        for a single text\protect\footnote{Data: The complete historical romances of Georg Ebers.}
        from the English eBooks collection.
        Idealization is defined by a pure power law of scaling $1-N/M$ 
        (red dashed line, see Materials and Methods).
        (\textbf{B}) The mixtures of all texts (gray points) and their idealizations (black points)
        from the English eBooks collection.
        Note that neither mixture results in a pure power law such as Zipf's ($\theta=1$, red, dashed line).
  }
  \label{fig:idealizedDists}
\end{figure}

As we have described, 
the CNCV model offers a means by which one can obtain a second scaling. 
The model is, like Simon's, 
framed as a model of the generation of a vocabulary. 
However, we are led to question whether lower
scalings are a product of vocabulary generation 
or an artifact of an interaction between disparate texts.
Suppose a collection of texts, $\mathcal{C} = \{T_1,...,T_{\numberoftexts}\}$, is read sequentially, 
and that each has rank-frequency distribution of Zipf/Simon form. 
Upon constructing idealized rank-frequency distributions 
from empirical data (see Materials and Methods),
we find that their combined distribution,
possesses multiple scaling regimes (see Fig.~\ref{fig:idealizedDists}).
Though each individual vocabulary might have been 
created without a decay of word introduction, 
an overlap in the words they use has it \textit{seem} as though the appearance 
of new words is rarer by the time the later texts are read. 
If one reads the texts repeatedly and in permuted orders, 
the resulting decay in the rate of word introduction 
likely does not evince itself until the mean text size 
(mean number of unique words per text) is reached, 
but certainly not before the minimum text size is reached.

\begin{figure}[t!]
  \includegraphics[width=1\columnwidth, angle=0]{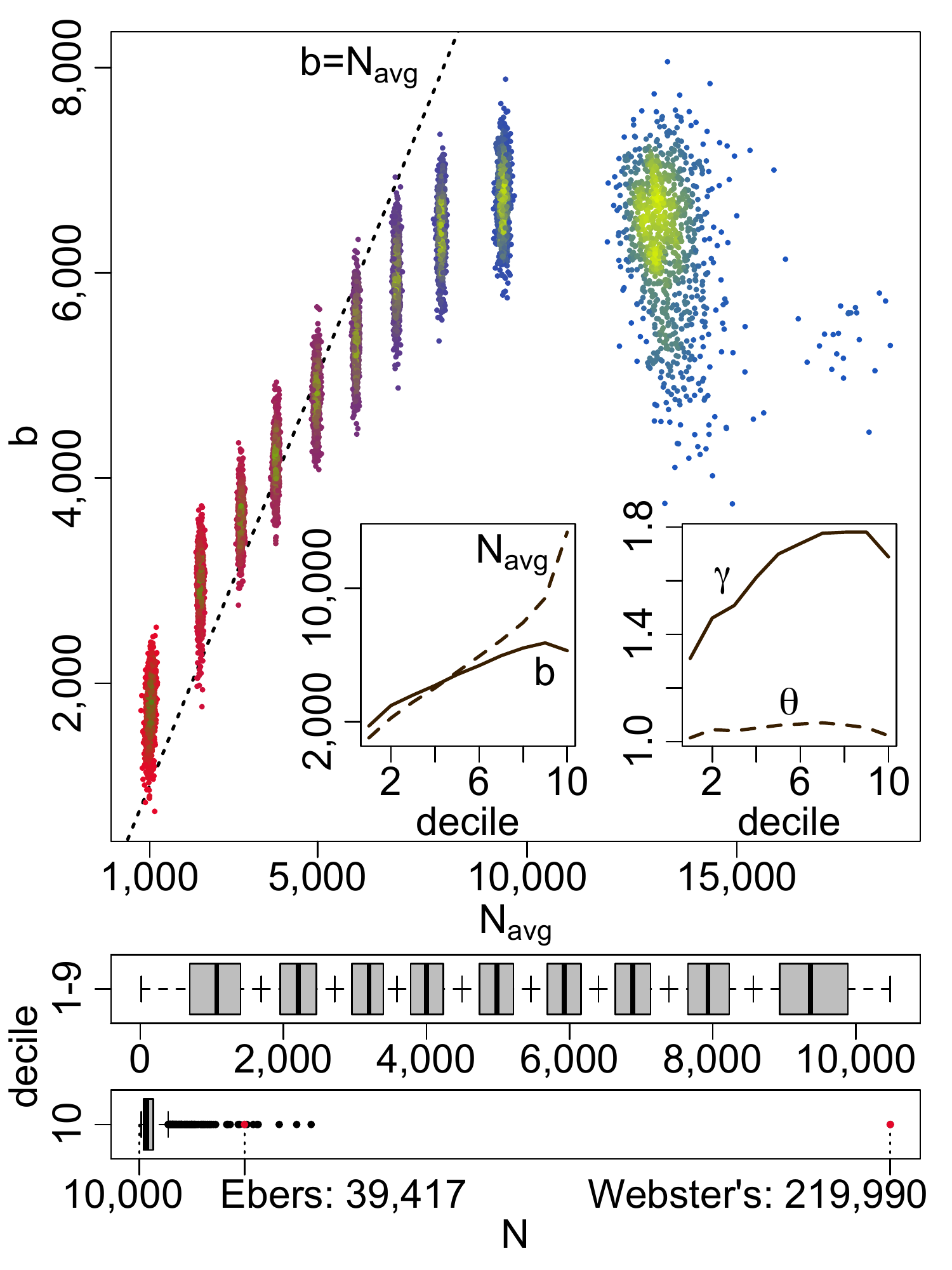}
  \caption{
    (\textbf{Top})~For each of the $10$ deciles of the English distribution of text sizes,
    we measure the parameters $b$, $\gamma$, $N_\text{avg}$, and $\theta$
    from $50$-book sample corpora. Each cloud represents $1,000$ sample corpora from 
    deciles $1$--$10$ 
    (low-to-high from left to right, where
    red to blue also indicates increasing decile
    and fade to green or yellow indicates increasing density).
    The line $b=N_\text{avg}$ is also presented
    (dashed line, main axis),
    and shows that $b$ increases with decile for all but the most extreme ($10^\text{th}$) decile.
    Main axes insets show parameter variation across deciles for both $b$ and $N_\text{avg}$ (left);
    and $\gamma$ and $\theta$ (right), where we note that Zipf's parameter, $\theta$, 
    is the only one that exhibits signs of stationarity.
    (\textbf{Bottom})~Box plots providing a more detailed look at the ten deciles of the distribution of text sizes.
    For clarity we have separated the plots for deciles $1$--$9$ from the $10^\text{th}$.
    This highlights the extreme nature of the later deciles (most notably the $10^\text{th}$),
    where the presence of poorly refined texts throw off estimates of $N_\text{avg}$,
    which we also note corresponds to the roll over in the distributions off of the $b=N_\text{avg}$
    axis above.
  }
  \label{fig:brkVavg}
\end{figure}

Operating under this ansatz---that a text mixing-derived 
scaling break, $b$, covaries with the mean number of unique words per text,
$N_\text{avg}$, in a corpus---we investigate thousands of corpora defined
by samples from the English eBooks database
(see Materials and Methods for more details on text sampling
 and a complete description of the eBooks database).
Obtaining $1,000$ text-sample corpora from each of the $10$ deciles of the
text-size distribution, we regress for $b$ (see Materials and Methods),
and record $N_\text{avg}$ to find that the two covary strongly
along the line $b=N_\text{avg}$ for all but the most extreme deciles
(see main axes Fig.~\ref{fig:brkVavg}, 
which we return to later in the discussion).
We see that this relationship breaks down 
in the presence of large-$N$ texts,
which upon closer inspection appear ill formed
in the sense of being of mixed origin themselves
(e.g., posthumous/longitudinal compendia, dictionaries, encyclopedias, etc...;
 see Materials and Methods~and~Fig.~\ref{fig:decayAndDistByWord_ebers} 
 for more details on corpus formation and internally-mixed texts).
Additionally, we see from these preliminary experiments
that both of the quantities, $b$ and $\gamma$,
do not appear as universal for a given language
(see Fig.~\ref{fig:brkVavg}),
but rather depend quite severely on corpus composition.
In fact, the only regressed parameter that presents
any signs of universality for a language is Zipf's 
exponent, $\theta$, which remains quite close to $1$.
These initial results indicate that hypotheses of
the locations of scaling breaks, $b$, corresponding
to language-universal lexical-core sizes
are in strong need of reevaluation,
or should be reformulated as corpus-relative.

\begin{examp}{}
  \begin{center}

  \end{center}
  Consider the two excerpts from Charles Dickens'\\
  ``A Tale of Two Cities'', taken as texts:
  \begin{align*}
    T_1:(&it,was,the,best,of,times,\\
    &it,was,the,worst,of,times),\hspace{10pt}\text{and}\\
    T_2:(&it,was,the,age,of,wisdom,\\
    &it,was,the,age,of,foolishness)
  \end{align*}
  Supposing we read $T_1$ first, the sequence of words is:
  \begin{align*}
    (T_1,T_2):(&{\color{red}it},{\color{red}was},{\color{red}the},{\color{red}best},{\color{red}of},{\color{red}times},\\
    &it,was,the,{\color{red}worst},of,times,\\
    &it,was,the,{\color{red}age},of,{\color{red}wisdom},\\
    &it,was,the,age,of,{\color{red}foolishness})
  \end{align*}
  where we have highlighted initial (growing text) word appearances in red.
  The corresponding sequences of values, $m,n_m,N_m,\alpha_m,A_m$ and $\alpha_m/A_m$, are then
  \begin{align*}
    m:(&{\color{red}1},{\color{red}2},{\color{red}3},{\color{red}4},{\color{red}5},{\color{red}6},7,8,9,{\color{red}10},11,12,\\
    &13,14,15,{\color{red}16},17,{\color{red}18},19,20,21,22,23,{\color{red}24})\\   
    n_m:(&{\color{red}1},{\color{red}2},{\color{red}3},{\color{red}4},{\color{red}5},{\color{red}6},6,6,6,{\color{red}7},7,7,\\
    &7,7,7,{\color{red}8},8,{\color{red}9},9,9,9,9,9,{\color{red}10})\\
    N_m:(&{\color{red}1},{\color{red}2},{\color{red}3},{\color{red}4},{\color{red}5},{\color{red}6},6,6,6,{\color{red}7},7,7,\\
    &8,9,10,{\color{red}11},12,{\color{red}13},13,13,13,13,13,{\color{red}14})\\
    \alpha_m:(&{\color{red}1},{\color{red}1},{\color{red}1},{\color{red}1},{\color{red}1},{\color{red}1},\frac{6}{7},\frac{6}{8},\frac{6}{9},{\color{red}\frac{7}{10}},\frac{7}{11},\frac{7}{12},\\
    &\frac{7}{13},\frac{7}{14},\frac{7}{15},{\color{red}\frac{8}{16}},\frac{8}{17},{\color{red}\frac{9}{18}},\frac{9}{19},\frac{9}{20},\frac{9}{21},\frac{9}{22},\frac{9}{23},{\color{red}\frac{10}{24}})\\     
    A_m:(&{\color{red}1},{\color{red}1},{\color{red}1},{\color{red}1},{\color{red}1},{\color{red}1},\frac{6}{7},\frac{6}{8},\frac{6}{9},{\color{red}\frac{7}{10}},\frac{7}{11},\frac{7}{12},\\
    &\frac{8}{13},\frac{9}{14},\frac{10}{15},{\color{red}\frac{11}{16}},\frac{12}{17},{\color{red}\frac{13}{18}},\frac{13}{19},\frac{13}{20},\frac{13}{21},\frac{13}{22},\frac{13}{23},{\color{red}\frac{14}{24}})\\
    \frac{\alpha_m}{A_m}:(&{\color{red}1},{\color{red}1},{\color{red}1},{\color{red}1},{\color{red}1},{\color{red}1},1,1,1,{\color{red}1},1,1,\\
    &\frac{7}{8},\frac{7}{9},\frac{7}{10},{\color{red}\frac{8}{11}},\frac{8}{12},{\color{red}\frac{9}{13}},\frac{9}{13},\frac{9}{13},\frac{9}{13},\frac{9}{13},\frac{9}{13},{\color{red}\frac{10}{14}}).
  \end{align*}  
  \caption{
    A concrete example of the text mixing effect,
    where we consider two passages ($T_1$ and $T_2$)
    as separate texts that are then mixed.
    The similarity of word use between these excerpts
    provides an excellent example for understanding the differences
    between the growing text, 
    where we count new word appearances ($n_m$)
    with the awareness of previous texts, 
    and the memoryless text,
    where we count word appearances ($N_m$) as new
    with each initial appearance in each text.
    Note that both $\alpha_m$ and $A_m$ are simply the
    quotients of $n_m$ and $N_m$ with $m$ (respectively),
    and that their quotient ($\alpha_m/A_m$) is equivalent
    to $n/N$, and is not equal to $1$
    \emph{only} when texts are mixed.
    \protect\label{ex:mixingExample}
  }
\end{examp}

In the following, we run text mixing experiments that measure decay in rates of word introduction
directly attributable to mixing texts to predict lower scalings in composite distributions.
As we read out texts (in some order) let $m$ be the volume of words observed at any point, and
$\memnum_m$ be the number of distinct words in the volume $m$, which we will refer to as the 
vocabulary size of the \emph{growing} text. To exhibit the effects of text mixing we contrast the vocabulary size of the 
growing text with the vocabulary size of the \emph{memoryless} text, $\memlessnum_m$, where we 
``forget'' the words read in all previous texts and continuing counting appearances of words that 
were initial \emph{in their text} (regardless of appearances in previous texts). 
From $\memnum_m$ and $\memlessnum_m$ we then have two proxies 
for the word introduction rate, one for the growing text $\alpha_m = \memnum_m/m$ and one for the 
memoryless text $A_m = \memlessnum_m/m$. We may consider $\alpha_m$ to be the word introduction rate 
of the composite (which includes mixing effects), and $A_m$ to be the word introduction rate of 
the individual texts (excluding mixing effects).

There are many conceivable mechanisms that lead to 
a power-law decay in the rate of word introduction. 
To measure the severity of scaling breaks we do not need to know
the true values of the word introduction rates, 
but instead just their scalings. 
So, to determine the 
extent to which text mixing generates word introduction decay, 
we isolate the portion of the scaling that 
results from mixing by measuring $\alpha_m/A_m$, 
the portion of word introduction remaining after mixing texts. 
Note that since $n_m\leq N_m$, 
one has $\alpha_m\leq A_m$, 
and hence $\alpha_m/A_m \leq 1$ for all $m$.
Hence, this normalized rate 
behaves as a non-constant only when mixing ensues, 
and so any decay measured via 
$\alpha_m/A_m$ implies the presence 
and is the direct consequence of text mixing
(see Example~\ref{ex:mixingExample} for an 
 intuitive understanding of all text mixing quantities).
Since $\alpha_m/A_m$ will be the only quantity 
used in the measurement of word introduction decay, 
we relax the notation, 
and simply write $\alpha$ for $\alpha_m/A_m$ 
and $\memnum$ for $\memnum_m$ in what follows.

To test the effects of text mixing, 
we not only observe the word introduction rate $\alpha(n)$, 
but consider its ability to predict 
the scalings of rank-frequency distributions. 
To do this, we note that by design, 
the data for $\alpha(n)$ are 
aligned with $f(r)$---both have domain $\{1,...,N_\text{corp}\}$ 
(where $N_\text{corp}$ is the vocabulary size of the corpus).
Further, since the theory has $\gamma = \theta + \mu$, 
we may also observe that $\alpha(\memnum)\cdot\memnum^{-\simonalpha}$,
need only be normalized
\begin{equation}
  \hat{p}(\memnum)=
  \frac{\alpha(\memnum)\cdot\memnum^{-\simonalpha}}{C}
  \text{, where }\\
  C=\sum_1^{N_\text{corp}}\alpha(\memnum)\cdot\memnum^{-\simonalpha}
\end{equation}
to produce a model for the normalized rank-frequency distribution 
$p(r)=f(r)/\sum_1^{N_\text{corp}}f(r)$. 
To determine a model's Zipf scaling, $\theta$, 
we scan the range $\{0.75,0.751,...,1.25\}$
and accept the $\theta$ for which $\hat{p}$
minimizes the sum of squares error 
\begin{equation}
  \sum_1^{N_\text{corp}}(\log_{10}p(r)-\log_{10}\hat{p}(r))^2
\end{equation}
over as many as $10,000$ $\log$-spaced ranks.

\begin{table}[b!]
  \begin{tabular}{|c|c|c|c|c|c|c|c|}
    \hline
     & $N_\text{books}$ & $N_\text{char}$ & $N_\text{min}$ & $N_\text{ave}$ & $b$ & $N_\text{max}$ & $N_\text{corp}$ \\\hline
    en & 19,793 & 46 & 5 & 5,899.3 & 5,849 & 219,990 & 2,836,900 \\\hline
    fr & 1,360 & 44 & 395 & 8,300.7 & 17,715 & 26,171 & 528,314 \\\hline
    fi & 505 & 31 & 1,144 & 8,872.6 & 7,761 & 31,623 & 811,742 \\\hline
    nl & 434 & 48 & 133 & 6,747.1 & 6,098 & 82,246 & 443,816 \\\hline
    pt & 375 & 38 & 203 & 4,675.8 & 10,363 & 17,818 & 246,497 \\\hline
    de & 327 & 30 & 153 & 7,554.9 & 7,259 & 113,089 & 477,274 \\\hline
    es & 223 & 34 & 406 & 8,735.1 & 15,079 & 29,452 & 237,874 \\\hline
    it & 194 & 29 & 1,083 & 9,388.7 & 13,954 & 29,445 & 258,509 \\\hline
    sv & 56 & 34 & 1,389 & 7,499.8 & 5,315 & 18,726 & 123,806 \\\hline
    el & 42 & 35 & 2,047 & 6,414.7 & 7,613 & 17,774 & 110,940 \\\hline
  \end{tabular}
  \caption{
    Table of information concerning the data used from the eBooks database.
    For each language we record the number of books ($N_\text{books}$); 
    the number of characters ($N_\text{char}$), which we take to be the number of letters~\cite{wikilatin,wikigreek} 
    (including diacritics and ligatures); 
    the minimum text size ($N_\text{min}$); 
    the maximum text size ($N_\text{max}$); 
    and the total corpus size ($N_\text{corp}$).
    For reference, we additionally record the regressed point of scaling break, $b$.
  }
  \label{tab:bookStats}
\end{table}

\section{Materials and methods} 
\label{sec:matmet}

In our experiments we work with a subset of the eBooks~\cite{gutenberg2012a} collection.
We collected those texts which were annotated sufficiently well to allow for the removal of meta-data 
as well as for the parsing of authorship, title, and language.
All together, this resulted in the inclusion of $23,309$ books from across ten languages 
(broken down in Tab. \ref{tab:bookStats}).

To idealize texts as discussed in Fig.~\ref{fig:idealizedDists} we note
that a resultant rank-frequency distribution from a pure Simon model
of constant word introduction rate, $\alpha_0$, will scale with Zipf exponent
$\theta=1-\alpha_0$, such that $N/M\rightarrow\alpha_0$ as the text grows.
Therefore, for an observed text of size $N$ and volume $M$, we define the
idealized Zipf/Simon exponent as $\theta_0=1-N/M$, 
and apply $\theta_0$ to the collection of ranks, $r=1,\cdots,N$, as
\begin{equation}
  f_\text{ideal}(r) = \left\lfloor \left(\frac{r}{N}\right)^{-\theta_0} +\frac{1}{2}\right\rfloor,
\end{equation}
while preserving their word-labels from the empirical data.

For all of the rank-frequency distributions analyzed,
we regress over as many as $10,000$ $\log$-spaced ranks
(taken over the range $r=1,...,N$) 
to determine estimates for $\theta$, $b$, and $\gamma$.
This estimation is done by applying a two-line least-squares regression,
constrained by intersection at the point of scaling break.
Given data points $(x,y)$, and a point of break, $x_b$,
we solve for the model
\begin{equation}
  \hat{y} = \left\{
  \begin{array}{lr}
    \beta_1 + \beta_2x, & : x \leq x_b\\
    \beta_3 + \beta_4x, & : x > x_b
  \end{array}
  \right.,
\end{equation}
constrained by $\beta_1 + \beta_2x_b = \beta_3 + \beta_4x_b$,
through standard minimization of the sum of squares error.
We compute this regression for $1,000$ $\log$-spaced points, $x_b$, across 
the middle $20$--$80\%$ of the $\log~r$ domain.
For given distribution we then perform these $1,000$ regressions
and accept the value $b$ for which we have observed the smallest SSE.

To understand our text mixing results we must note
that there is measurement error for both $b$ and $N_\text{avg}$. 
As a regressed quantity, this may be expected for $b$,
but for $N_\text{avg}$, the existence of measurement error is less obvious,
and generally results from poor corpus composition. 
The main effect stems from the fact that 
many texts in the eBooks data set are internally mixed.
The longitudinal compendia of individual authors and genres are
the most intuitive and abundant examples of internally mixed texts,
and the most extreme cases are generally reference texts,
e.g., dictionaries, encyclopedias, and textbooks
(see Fig.~\ref{fig:brkVavg}).
The major point is that when a compendium is not refined, 
but taken as an individual text in a corpus, 
the calculation of $N_\text{avg}$ 
considers only a single book of large size (wrongly),
instead of many books of smaller size (correctly).
Within the English data set we have found that
the large-$N$ texts are generally 
of this variety and dominate the $10^\text{th}$ decile.
Reading down the top ten $N$-ranking texts 
makes this abundantly clear:
\begin{itemize}
\item[1.] Webster's Unabridged Dictionary
\item[2.] Diccionario Ingles-Espa\~{n}ol-Tagalog
\item[3.] The Complete Project Gutenberg Works of George Meredith
\item[4.] The Anatomy of Melancholy
\item[5.] A Concise Dictionary of Middle English
\item[6.] A Pocket Dictionary
\item[7.] The Nuttall Encyclopaedia
\item[8.] The Complete PG Works of Oliver Wendell Holmes, Sr.
\item[9.] The Complete Historical Romances of Georg Ebers
\item[10.] The Complete Project Gutenberg Works of Galsworthy
\end{itemize}
Note here that among these compendia and reference
texts lies a two way (Spanish/English) dictionary
whose placement in the top $10$ likely results
from dual word forms (English and Spanish translations) 
of the majority of words that it possesses.
We have explored the impact of these under-refined 
and ill-formed texts in detail in Fig.~\ref{fig:brkVavg}, 
where we have found a clear association of $b$ with $N_\text{avg}$
along the line $b=N_\text{avg}$
that breaks down in the larger deciles,
where these strange texts occur.

\begin{figure}[t!]
  \centering
  \includegraphics[width=0.99\columnwidth, angle=0]{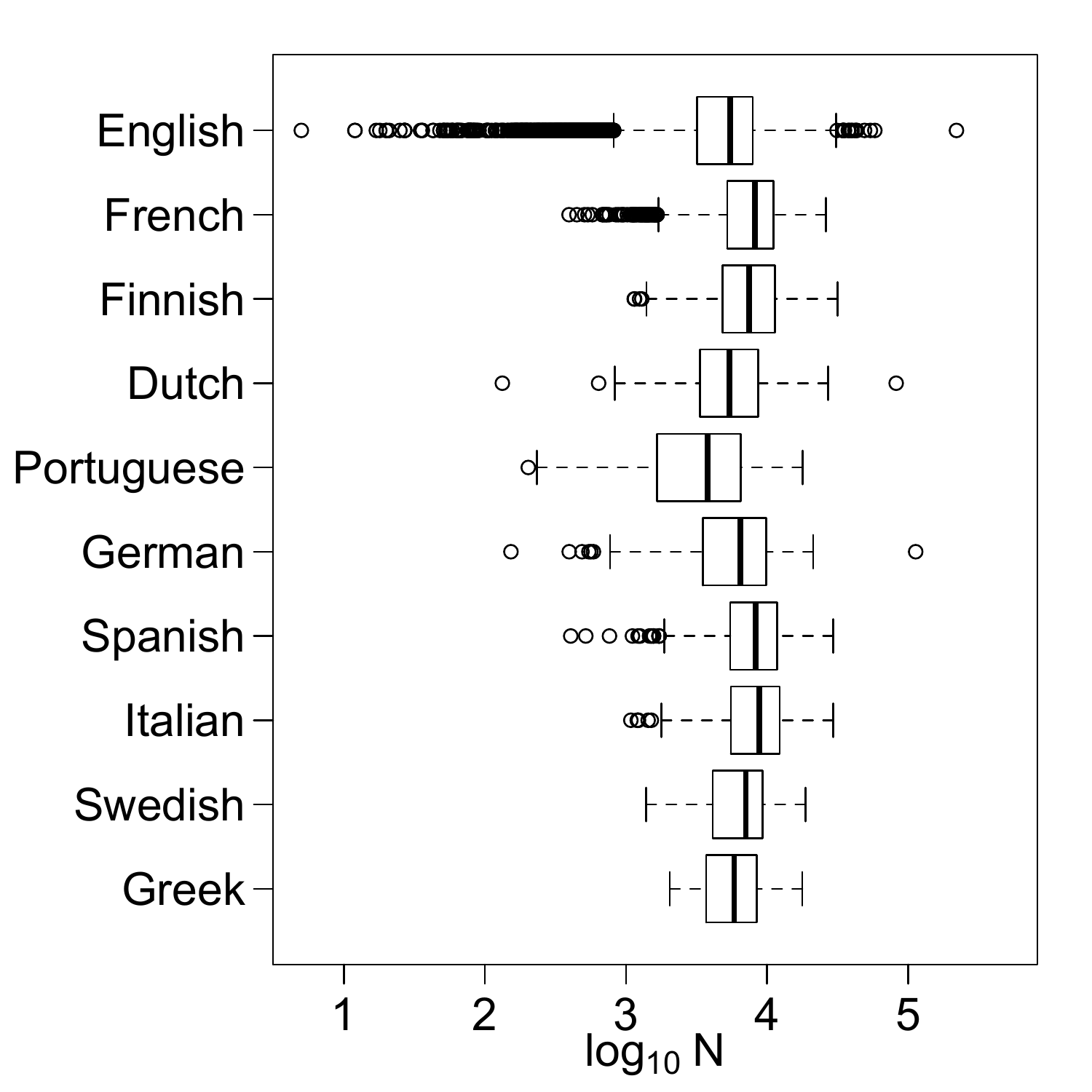}
  \caption{
    Box plots of the base ten logarithm vocabulary sizes of the texts contained in the 
    $10$ eBooks corpora studied. 
    Center bars indicate means
    and whiskers extend to most extremal values up to $1.5$ times the I.Q.R. length,
    whereupon more extremal values are plotted as points designated ``outliers''.
  }
  \label{fig:languagesBoxPlot}
\end{figure}

We also note that $N_\text{avg}$ is subject to measurement error 
from overrefined texts as well, 
most notably in the Portuguese data set,
which has the smallest average text size,
while having the fifth largest number of books
(see Tab.~\ref{tab:bookStats} and Fig.~\ref{fig:languagesBoxPlot}).
There we note that Portuguese presents 
the most significant deviation between
$N_\text{avg}$ and $b$
($b$ is notably more than $120\%$ larger than $N_\text{avg}$), 
and moreover that this deviation is in the
expected direction, i.e., $N_\text{avg}\ll b$.
Note also that this observation is in 
agreement with those other languages
that have $N_\text{avg}\ll b$ in Tab.~\ref{tab:bookStats}
(specifically Italian, Spanish, and French),
where in Fig.~\ref{fig:languagesBoxPlot}
we see that having many low-$N$ outliers
with no high-$N$ outliers 
biases the corpus-wide measurement of $N_\text{avg}$.

To estimate $\mu$ we perform 
common least squares linear regression 
on the $\log$-transformed data 
over the region $[N_\text{avg},N_\text{corp}]$,
since $N_\text{avg}$ is generally the point 
at which mixing-derived decay becomes clear.

Computation of $\alpha(\memnum)$ involves 
running many realizations of the text mixing
procedure, randomizing the order in which the texts are read. 
To ensure that our measurements are accurate, 
we adhere to a heuristic---that the number of text mixing runs be no less than
$10\cdot N_\text{books}$ for the given corpus.
The final values used is in our experiments 
are computed as averages of the $\alpha_m/A_m$
from the more than $10\cdot N_\text{books}$ runs. 
However, we note that $\alpha_m/A_m=\memnum_m/N_m$, 
where $\memnum_m$ ranges with rank: $n_m=1,2,3,\cdots,N_\text{corp}$. 
So, the only quantities that vary across runs
that are necessary to compute $\alpha(\memnum)$ are the $N_m$. 
Hence we take the average as $\alpha(\memnum_m) = \memnum_m/\langle N_m \rangle$, 
which is in fact the harmonic mean of the $\alpha(\memnum_m)$ 
(the truest mean for rates).

In our investigation of the different divisions of
the internally mixed corpus,
``The complete historical romances of Georg Ebers,''
we have shown how important it is to have meaningfully
defined texts to be able to produce an accurate
text mixing model for a corpus. 
An important component of this exhibition 
presented the extremal refinement, 
where each word is treated individually as a separate text
(a highly non-realistic scenario).
To conduct a text mixing experiment for such a refinement
can be quite computationally taxing,
as this requires taking permutations 
of the word orders of the entire corpus.
Since this process is entirely independent
of the original word orderings from the corpus,
it may be computed directly from the
rank-frequency distribution via expected gap sizes.
In particular, we wish to determine the average number 
of previously seen words appearing between the
$n^\text{th}$ and $n+1^\text{st}$ ``new'' words,
given all permutations of the corpus words.
Denoting this number by $\overline{M}_n$,
we note that the average word introduction rate over this range
is easily found as $\alpha_n=1/\overline{M}_n$.
We then define $i_n$ as 
the total number of previously-observed words 
that were not yet counted 
by the time the $n^\text{th}$ new word was observed, 
and define $j_n$ to be
the total number (out of all corpus words) 
that were not yet counted 
by the time the $n^\text{th}$ new word was first observed
(including those word types that were not yet observed). 
Then, if $P_n(M)$ is the
probability that the $n^\text{th}$ and $n+1^\text{st}$ 
``new'' words were separated by precisely $M$ previously seen words,
\begin{equation}
  \begin{split}
    \overline{M}_n &= \sum_{M=0}^{i_n}M\cdot P_n(M)\\
    &=\sum_{M=0}^{i_n}M\cdot\frac{j_n-i_n}{j_n-M}\prod_{k=0}^{M-1}\frac{i_n-k}{j_n-k}
  \end{split}
\end{equation}
where in the last expression,
the product is the probability of 
seeing $M$ consecutive previously-observed words, 
with the first factor being 
the probability that the ``new'' word is seen 
as the $M+1^\text{st}$. 
These expressions for the $\overline{M}_n$ 
are iteratively computable, 
and in addition, 
since the sums converge quickly, 
we find that it suffices to 
take their first $1,000$ terms 
for added computational efficiency.

\begin{figure}[t!]
  \centering
  \includegraphics[width=0.99\columnwidth, angle=0]{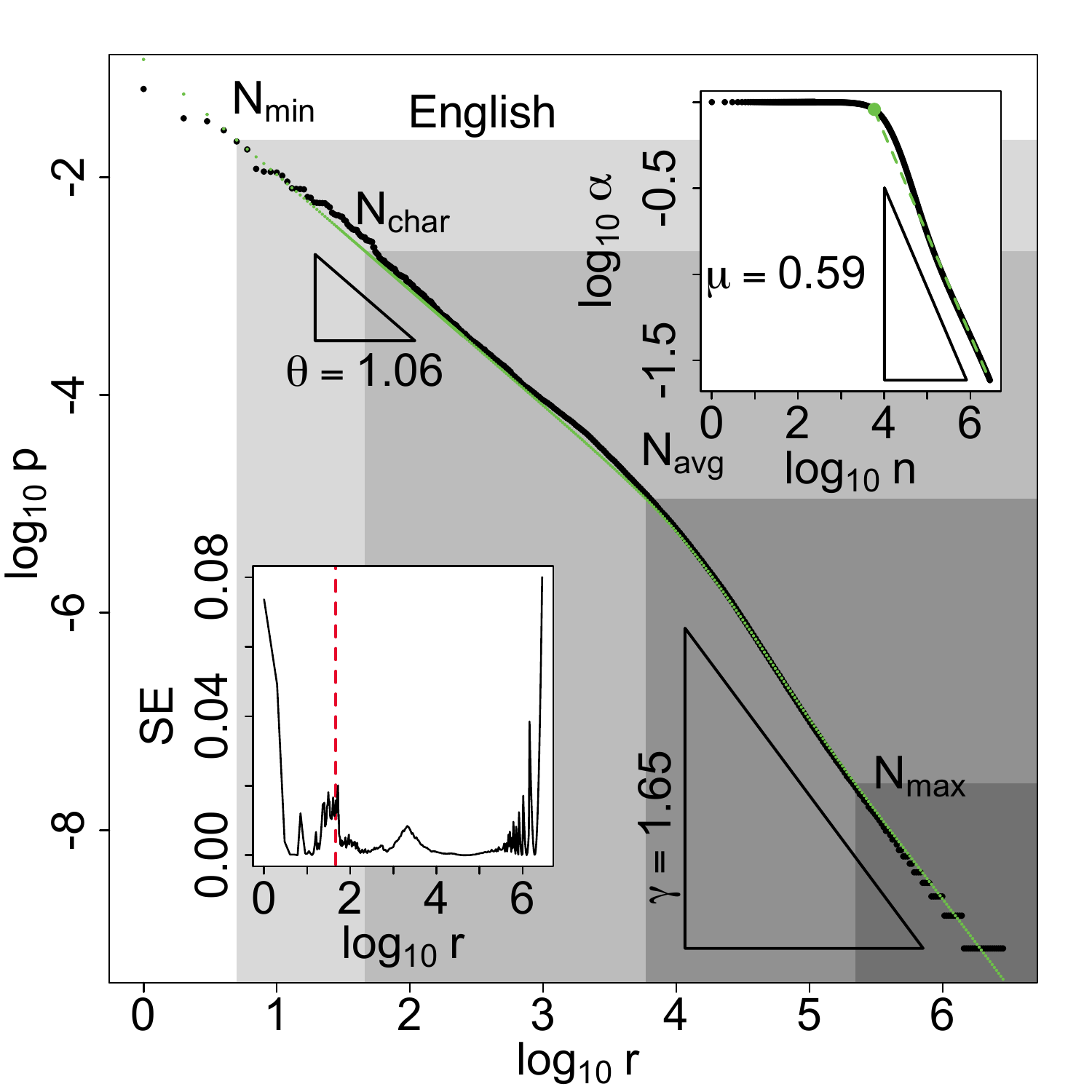}
  \caption{    
    The results of text mixing experiments for the largest, English corpus from the eBooks collection.
    The main axes show the empirical, normalized rank-frequency distribution (black points), $p$, 
    and the model determined by text mixing (green points), $\hat{p}$.  
    The measured lower and upper exponents, $\gamma$ and $\simonalpha$, 
    are depicted in the lower-right and upper-left respectively,
    with triangles indicating the measured slopes.    
    We also present gray boxes in the main axes to highlight the different mixing regimes, 
    marked by $N_\text{char}$, $\Nmi$, $\Nav$, and $\Nma$
    (see Materials and Methods and Tab.~\ref{tab:bookStats} 
    for complete descriptions of these quantities).
    The lower left inset shows the point-wise squared error $(p(r)-\hat{p}(r))^2$,
    whose sum is minimized in the transformation of $\alpha$ into $\hat{p}$.
    The upper right inset shows the untransformed rate of word introduction, 
    $\alpha$ (black points), 
    and the decay exponent $\mu$, 
    which is depicted by the regressed slope (green dashed line).
  }
  \label{fig:decayAndDistByWord_en}
\end{figure}

\begin{figure*}[t!]
  \centering
  \includegraphics[width=0.99\textwidth, angle=0]{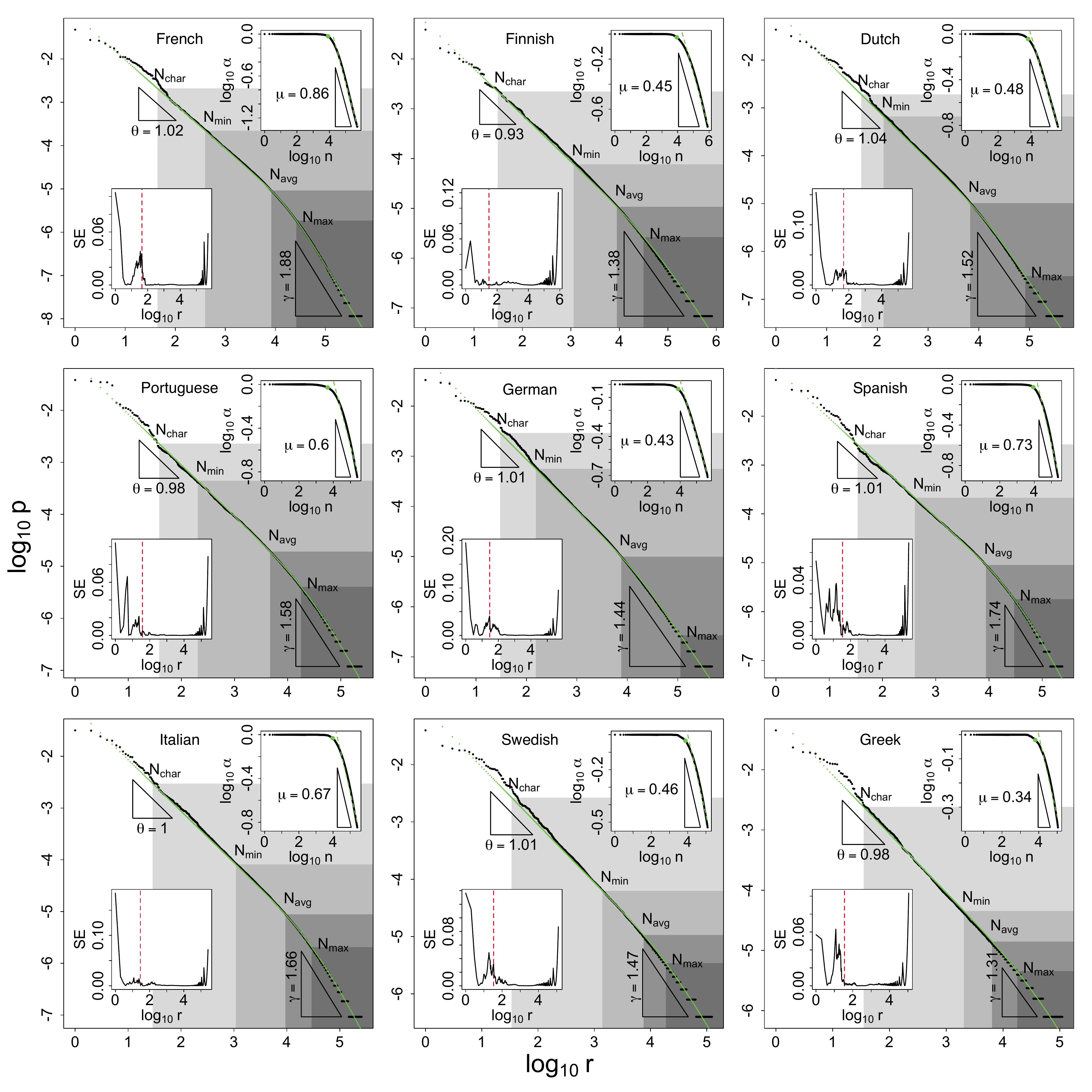}
  \caption{
    The results of text mixing experiments for the nine smaller corpora analyzed.
    All insets, color-coding, and labels are consistent with those from
    the larger, English presentation in Fig.~\ref{fig:decayAndDistByWord_en}, 
    whose caption possesses full descriptions of all axes and plotted data.
  }
  \label{fig:decayAndDistByWord}
\end{figure*}

\begin{figure*}[t!]
  \centering
  \includegraphics[width=0.99\textwidth, angle=0]{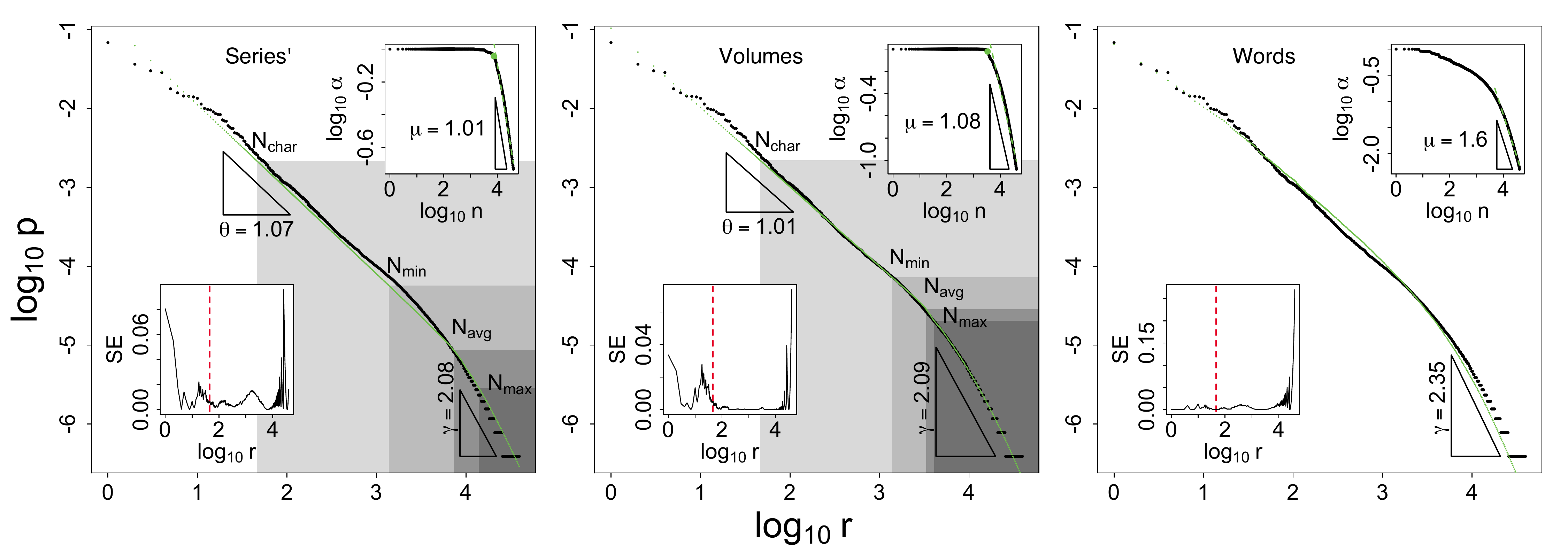}
  \caption{
        Text mixing results for a single-author corpus. 
        Here, $\alpha$ was measured for differing refinements of the 
        Egyptological fiction compendium/text 
        ``The complete historical romances of Georg Ebers'' into sub-texts. 
        All insets, color-coding, and labels are consistent with those from
        the English presentation in Fig.~\ref{fig:decayAndDistByWord_en}, 
        whose caption possesses full descriptions of all axes and plotted data.
        \textbf{(Left)} Each series is considered a separate text.
        \textbf{(Middle)} Each volume of each series is considered a separate text.
        \textbf{(Right)} Each word (the extremal refinement, see Materials and Methods) 
        in the compendium is considered a separate text.
        Note that in the upper right insets,
        $\alpha$ decreases overall with each refinement
        (as by definition it must),
        and that there appears to be an optimal refinement 
        for producing a text mixing model,
        likely close to the scale of volumes.
  }
  \label{fig:decayAndDistByWord_ebers}
\end{figure*}

\section{Results and discussion}

To understand our results we define 
$\Nmi$, $\Nav$ and $\Nma$ as the 
minimum, average, and maximum text sizes 
(by numbers of unique words) respectively
(see Tab.~\ref{tab:bookStats}).
These three values obviate four text mixing regimes:
\begin{align*}
  n<\Nmi &\text{;\: Zipf/Simon (no mixing)}\\
  \Nmi\leq n \leq \Nav &\text{;\: initial (minimal mixing)}\\
  \Nav\leq n \leq \Nma &\text{;\: crossover (partial mixing)}\\
  n>\Nma &\text{;\: terminal (full mixing)}
\end{align*}
In the Zipf/Simon regime 
we expect the result of an unperturbed Simon model, 
though because mixing is also 
minimal over the initial regime 
we expect that behavior over the first two regimes 
to more or less be consistent.
Once in the crossover regime, 
words will on average have appeared under the effects of text mixing
and so there is the expectation that 
$\Nav$ will mark the macroscopically observable change in behavior, 
or scaling break of the rank-frequency distribution, i.e., we expect $b\approx\Nav$. 
Plotting the two against one another, 
we have see this relationship holds across 
sample corpora from the well-behaved deciles
of the English distribution of text sizes
(see Fig.~\ref{fig:brkVavg}),
and breaks down in the presence
of ill-formed texts.
Finally, over the terminal regime, 
all words will appear in the presence of mixing, 
and so this regime exhibits the stabilized
second scaling, characterized by the decay parameter $\mu$.

Our main results from text mixing,
comparing the text mixing-derived model, $\hat{p}$,
with the normalized empirical rank-frequency data, $p$,
may be found for the English data set in Fig.~\ref{fig:decayAndDistByWord_en},
and for the nine other languages studied in Fig.~\ref{fig:decayAndDistByWord}.
For all $10$ languages 
we observe that the models defined by text mixing, $\hat{p}$, 
produce excellent predictions of the 
rank-frequency distributions
(Main axes, Figs.~\ref{fig:decayAndDistByWord_en}~and~\ref{fig:decayAndDistByWord}),
which is made quite clear by 
plotting point-wise squared error
(lower-left insets, Figs.~\ref{fig:decayAndDistByWord_en}~and~\ref{fig:decayAndDistByWord}).
For each corpus 
we see a broad range of ranks
beginning not far before $10^2$,
and extending into the second scaling
where the error is quite low 
(disregarding the effect of the finite-size plateaux).

We also perform text mixing analysis at different scales
for a single, large, and internally mixed text from the English data set,
``The complete historical romances of Georg Ebers.''
It is important to note 
before interpreting these results
that the text itself is a compendium,
combining series' that were each 
written by the author 
over the course of more than $30$ years,
writing and publishing volumes independently.
With this in mind, 
the text offers an important
example for text mixing that
helps us to understand several important details.
First, that not all texts are well formed---an individual text 
such as this may in of itself present a scaling break
that has resulted from text mixing.
Second, that the scaling break of a single,
large text may be understood through text mixing analysis.
This second point is more difficult observe, 
as it requires an appropriate 
refinement of the internally-mixed text, 
i.e., one must be able to break the mixed text 
into appropriately independent sub-texts.
From our example in Fig.~\ref{fig:decayAndDistByWord_ebers},
we can see that the division of the text into a corpus
of 28 series' (left panel) renders a text mixing model for the
empirical data with much higher error than a division
into a corpus of $143$ volumes 
(center panel, a refinement of the series' division).
We also present text mixing results from the extremal refinement,
where each individual word is treated as a text 
(right panel, see Materials and Methods for 
 more information on the extremal refinement),
which shows that a text can be over-refined
to produce a poor text mixing model.

It is worth noting from our results that the parameter, $\simonalpha$,
is frequently measured to lie outside the Simon-productive range, $(0,1)$.
Therefore, we are left to conclude that individually, 
many texts are subject to internally-derived decay in word introduction rates
(as is exemplified by the Ebers text in Fig.~\ref{fig:decayAndDistByWord_ebers}),
i.e., the underlying rank-frequency distributions are not of pure Zipf/Simon form
(as we suggest in other work~\cite{williams2014a}),
but, instead, subject to internal mixing.
Though we do not exhaustively investigate the occurrence of 
internally-derived decay in the rates of word introduction
across the eBooks data set,
it seems quite possible that all of the texts parsed 
are subject to some internal mixing effects, 
whether from non-original annotation 
by the Project Gutenberg e-Text editors, 
or just the mixing of differing components 
(e.g., chapters, series', volumes, prologues, etc...).
This of course would require that 
these mixing effects be of low-impact
in the cases generally considered strong examples Zipf's law.

We also note a strange behavior 
(which is captured by the text mixing model) 
in the English data set.
There, we have found a relatively shallow lower scaling ($\gamma\approx 1.65$), 
but notice that it appears to be one of possibly two lower scalings. 
For English, the crossover regime exhibits 
a consistently steeper scaling 
that dies away in the terminal regime. 
Though we have no certain explanation for this behavior, 
part of what makes the English collection 
so different from the others is the sheer number of texts
(see Tab.~\ref{tab:bookStats}).
However, upon looking closer at the 
distribution of English text sizes,
we also notice that the collection
possess some extremely large-$N$ outliers,
In the largest text
(which has nearly an order of magnitude more words than any other text)., 
approximately one tenth of all words are represented
(out of nearly $20,000$ books),
which must have a profound impact on the combined
rank-frequency distribution, and hence lower scaling.
Further, this large-$N$ hypothesis is supported by 
our preliminary investigation (see Fig.~\ref{fig:brkVavg}) 
where we observed that those (large) texts in the tenth decile
not only generated scaling break points that went against
the $b=N_\text{avg}$ correspondence,
but also, generated relatively shallow lower scalings,
against the trend of steepening with increasing decile.
English is also well-known for its willingness to adopt foreign words, 
which may lead to an increased rate of appearance of low-count loan words.
Regardless of the reasons for this difference with English, 
we find that text mixing captures the shape of both lower scaling regimes, 
and so both are well explained by the text mixing model.

We also take time to make note of and discuss another 
anomalous behavior of the rank-frequency distributions investigated.
Upon viewing a rank-frequency distribution for Zipf's law, 
one generally finds a ``wobble'' of the frequency data around Zipf's scaling
(regardless of the existence of a scaling break).
We refer to the termination of this ``wobble''
as the point of stabilization of the Zipf/Simon regime.
Looking at the empirical data from the ten languages,
we see that this stabilization point generally appears
early on the in Zipf/Simon regime,
and generally not before the first $10^2$ ranks.
Though we have no definitive explanation for the
existence of this anomaly,
we note upon looking at the pointwise-squared errors
that the stabilization point frequently occurs
near each language's number of characters, $N_\text{char}$
(depicted as a red dotted vertical line in
 each of the lower left insets of 
 Figs.~\ref{fig:decayAndDistByWord_en},~\ref{fig:decayAndDistByWord},~and~the~center~panel~of~\ref{fig:decayAndDistByWord_ebers}).
Whether the numbers of characters spawned in the
generation of primordial, character-based languages
still influence the shapes of rank-frequency distributions
of descendant languages today,
we cannot say for sure.
However this anomalous regime 
appears consistently across languages,
and may potentially be of consistent shape 
across the corpora of a language.
If so, we might view such anomalies 
as universal properties of languages,
and so highlight them in the 
hopes of opening a broader discussion.

In light of the results presented, 
we take time to consider the validity 
of the core language hypothesis. 
We have seen significant variation 
in both the location and severity of scaling breaks 
both across and within languages. 
Upon sampling the English corpus by deciles,
we have observed that the regressed point of scaling break, $b$, is not stationary 
(see Fig.~\ref{fig:brkVavg}).
We take this as indication of 
the lack of validity of and language-universal core/non-core hypothesis,
as a core should exhibit a strong consistency of size. 
Moreover, languages closely related via a common, recent ancestor 
should likewise exhibit this consistency, 
but notably two of the languages most closely related in the study, 
Spanish and Portuguese, present a large difference in $b$, 
($10,363$ for Spanish, and $15,079$ for Portuguese---see Tab.~\ref{tab:bookStats}). 
Both of these results seem to indicate that 
scaling breaks in rank-frequency distributions
are likely consequences of text and corpus composition.
Hence, it may then be more reasonable to consider a language core 
as a collection of words necessary for basic description, 
but not overlapping in use or meaning.
However, such a core lexicon would 
need be determined by native practitioners, 
and not necessarily be an 
observable property of rank-frequency distributions. 
Alternatively, one could consider a corpus-core 
by it's collection of words common to it's texts. 
However, such a ``common core'' would be 
entirely dependent on the composition of the corpus,
and hence not a universal property of a language proper.

\clearpage

\clearpage

\newwrite\tempfile
\immediate\openout\tempfile=startsupp.txt
\immediate\write\tempfile{\thepage}
\immediate\closeout\tempfile

\setcounter{page}{1}
\renewcommand{\thepage}{S\arabic{page}}
\renewcommand{\thetable}{S\arabic{table}}
\setcounter{table}{0}

\end{document}